\documentclass{article} 
\usepackage{iclr_set,times,xspace}
\usepackage{multirow}
\usepackage{xcolor}
\usepackage{graphicx}
\usepackage{boldline}
\usepackage{booktabs}
\usepackage{amssymb}
\usepackage{algorithm}
\usepackage{algpseudocode}
\usepackage{wrapfig}
\usepackage{floatflt}

\usepackage{amsmath,amsfonts,bm}

\def\eqref#1{equation~\ref{#1}}

\def\1{\bm{1}}

\DeclareMathAlphabet{\mathsfit}{\encodingdefault}{\sfdefault}{m}{sl}
\SetMathAlphabet{\mathsfit}{bold}{\encodingdefault}{\sfdefault}{bx}{n}

\def\gD{{\mathcal{D}}}

\def\gL{{\mathcal{L}}}

\newcommand{\E}{\mathbb{E}}

\usepackage{hyperref}
\usepackage{url}
\newcommand{\attack}{{\sc Peta}\xspace}

\title{\attack: Parameter-Efficient Trojan Attacks}

\iclrfinalcopy

\author{Lauren Hong \thanks{Work done while visiting Stony Brook University} \\
Stony Brook University \\
\And
Ting Wang \\
Stony Brook University \\
}

\begin{document}

\maketitle

\begin{abstract}
Parameter-efficient fine-tuning (PEFT) enables efficient adaptation of pre-trained language models (PLMs) to specific tasks. By tuning only a minimal set of (extra) parameters, PEFT achieves performance that is comparable to standard fine-tuning. However, despite its prevalent use, the security implications of PEFT remain largely unexplored. In this paper, we take the initial steps and present \attack, a novel trojan attack that compromises the weights of PLMs by accounting for downstream adaptation through bilevel optimization: the upper-level objective embeds the backdoor into a model while the lower-level objective simulates PEFT to both retain the PLM's task-specific performance and ensure that the backdoor persists after fine-tuning. With extensive evaluation across a variety of downstream tasks and trigger designs, we demonstrate \attack's effectiveness in terms of both attack success rate and clean accuracy, even when the attacker does not have full knowledge of the victim user's training process.
\end{abstract}

\section{Introduction}

Backdoor attacks \citep{gu2017badnets}, also known as trojan attacks, are widely-studied training-time security threats to deep neural networks. In these scenarios, the attacker aims to inject a backdoor into a victim model such that the model behaves normally on benign inputs and gives attacker-specified outputs upon seeing examples that contain predefined triggers. In the context of natural language processing (NLP), attackers can achieve this by releasing poisoned datasets, compromised pre-trained language model (PLM) weights, or trojaned models that are intended to be used out of the box \citep{openbackdoor,kurita2020weight,EP,trojanlm,zhang2021neural,yang2021rethinking,qi2021mind,pan2022hidden}.

Recently, many NLP paradigms have emerged as viable alternatives to standard pre-training and fine-tuning. However, the unique characteristics of these paradigms introduce a myriad of unique vulnerabilities. For example, \citet{kandpal2023backdoor} designed a backdoor attack for in-context learning \citep{brown2020language}, a strategy for eliciting the ability to perform a desired task without requiring any updates to the model parameters. Additionally, \citet{mei-etal-2023-notable} and \citet{xu2022exploring} explore new possibilities in prompt-based learning, a paradigm that reformulates classification tasks into the cloze task, which is known to be effective for few-shot learning \citep{schick-schutze-2021-exploiting,gao-etal-2021-making}.

In this work, we focus on parameter-efficient fine-tuning (PEFT). Unlike the conventional fine-tuning paradigm that requires retraining all of the PLM's parameters, PEFT only fine-tunes a minimal set of (extra) parameters while keeping the PLM's original weights frozen \citep{adapter,prefix-tuning,prompt-tuning,lora}. It is shown that PEFT not only curtails the prohibitive training costs in terms of both data and compute resources but also achieves performance that is comparable to full-scale fine-tuning~\citep{petl-unify,prefix-tuning}.

Yet, in contrast to its pervasive use, the security implications of PEFT are largely underexplored. We take the initial steps in this line of research and present \attack\footnote{\attack: \underline{P}arameter-\underline{E}fficient \underline{T}rojan \underline{A}ttack}, a novel trojan attack tailored to PEFT, which consists of two stages: (1) \textbf{bilevel optimization}, in which the attacker inserts the backdoor into a general-purpose pre-trained language model  and (2) \textbf{parameter-efficient fine-tuning} on a clean dataset, which is performed by the victim user. 

\section{Methodology}

\textbf{Backdoor Attacks} - In the classification setting, an adversary who wants to launch a backdoor attack on some classifier $f(\cdot)$ has the following requirements for the model: (1) the classifier should output a target label $t$ whenever a trigger is inserted into an example and (2) the classifier should behave normally when given examples without triggers. More specifically, for any example $x$ with true label $y$, let $\hat{x}$ denote the poisoned version of $x$ (i.e., the result of inserting a trigger into $x$). The attacker hopes to manipulate the training process of $f(\cdot)$ such that $f(x) = y$ and $f(\hat{x}) = t$. For textual backdoor attacks, the triggers can be seemingly innocuous character patterns \citep{kurita2020weight,EP,trojanlm,zhang2021neural,yang2021rethinking}, sentences \citep{addsent,chen2021badnl}, writing styles \citep{pan2022hidden, qi2021mind}, or syntactic structures \citep{SynBkd}.

\begin{table}[t]
\begin{minipage}[t]{0.52\linewidth}
\caption{PEFT transfer results. In the \textbf{PEFT} column, the left side of the arrow is the proxy method employed during the first stage of \attack, while the right side is the method used by the victim user.}
\label{peft-table}
\begin{center}
\resizebox{\textwidth}{!}{
\begin{tabular}{ c | c |  c  c  c  c }
\toprule[0.1em] 
\multirow{2}{3em}{} & \multirow{2}{3em}{\textbf{PEFT}} & \multicolumn{2}{ c }{\textbf{Style}} & \multicolumn{2}{ c }{\textbf{Syntax}}  \\
& & ACC & LFR & ACC & LFR \\
\midrule[0.1em]
\multirow{2}{1.5em}{\textbf{OE}} & \textbf{L} $\rightarrow$ \textbf{A} & 85.22 &	92.73 &	84.87 &	99.84 \\
& \textbf{A} $\rightarrow$ \textbf{L} & 84.63 &	96.28 &	84.05 &	99.84 \\
\midrule[0.1em] 
\multirow{2}{1.5em}{\textbf{AG}} & \textbf{L} $\rightarrow$ \textbf{A} & 90.67 &	99.79 &	90.51 &	99.93 \\
& \textbf{A} $\rightarrow$ \textbf{L} & 91.29 &	99.86 &	90.87 &	99.91 \\
\bottomrule[0.1em]
\end{tabular}}
\end{center}
\end{minipage} 
\hfill  
\begin{minipage}[t]{0.435\linewidth}
\caption{Domain transferability results. For datasets X and Y, X $\to$ Y means that X was used to compromise the PLM's weights and Y was used by the victim user during PEFT.}
\label{domain-table}
\begin{center}
\resizebox{\textwidth}{!}{
\begin{tabular}{ c c c c c }
\toprule[0.1em] 
\multirow{2}{3em}{\textbf{Attack}} & \multicolumn{2}{ c }{\textbf{AG $\rightarrow$ TT}} & \multicolumn{2}{ c }{\textbf{TT $\rightarrow$ AG}}  \\
& ACC & LFR & ACC & LFR \\
\midrule[0.1em]
Clean & 87.36 &	0.56 &	89.97 &	4.42 \\
Upper-Only & 87.66 &	86.67 & 89.84 &	91.16 \\
LWP & \textbf{88.07} &	\textbf{100} &	89.82 &	90.94 \\
BadNet & 87.3 &	95.83 &	\textbf{90.22} &	95.32 \\
\attack & 87.12 &	\textbf{100} &	89.64 &	\textbf{98.79} \\
\bottomrule[0.1em]
\end{tabular}}
\end{center}
\end{minipage}
\end{table}

\textbf{Weight Poisoning Attacks on PEFT} - PEFT is an efficient alternative for adapting PLMs to specific tasks~\citep{petl-unify}. Given a frozen pre-trained language model $f(\cdot; \theta)$, PEFT methods insert additional parameters $\delta$ in $f(\cdot; \theta)$ to create a new function $\bar{f}(\cdot; \theta, \delta)$ and trains $\delta$ while keeping $\theta$ fixed. In previous work on inserting backdoors in the PEFT paradigm via weight poisoning, the attacker poisons the \textit{PEFT weights} and releases them to a victim user, who will use them for initialization to do further PEFT training \citet{gu-etal-2023-gradient}. In contrast, we consider a novel approach that targets the standard setting where the newly inserted PEFT parameters are randomly initialized during downstream adaptation and release compromised \textit{PLM weights} to the user instead, which will remain frozen during PEFT. Like in existing trojan attacks that embed backdoors in pre-trained language models for the regular fine-tuning paradigm \citep{kurita2020weight, por, li-etal-2021-backdoor,chen2021badpre}, the PLM weights in our attack should be poisoned such that the backdoor doesn't get overwritten after fine-tuning.

\begin{wrapfigure}{r}{0.48\textwidth}
\includegraphics[width=0.48\textwidth]{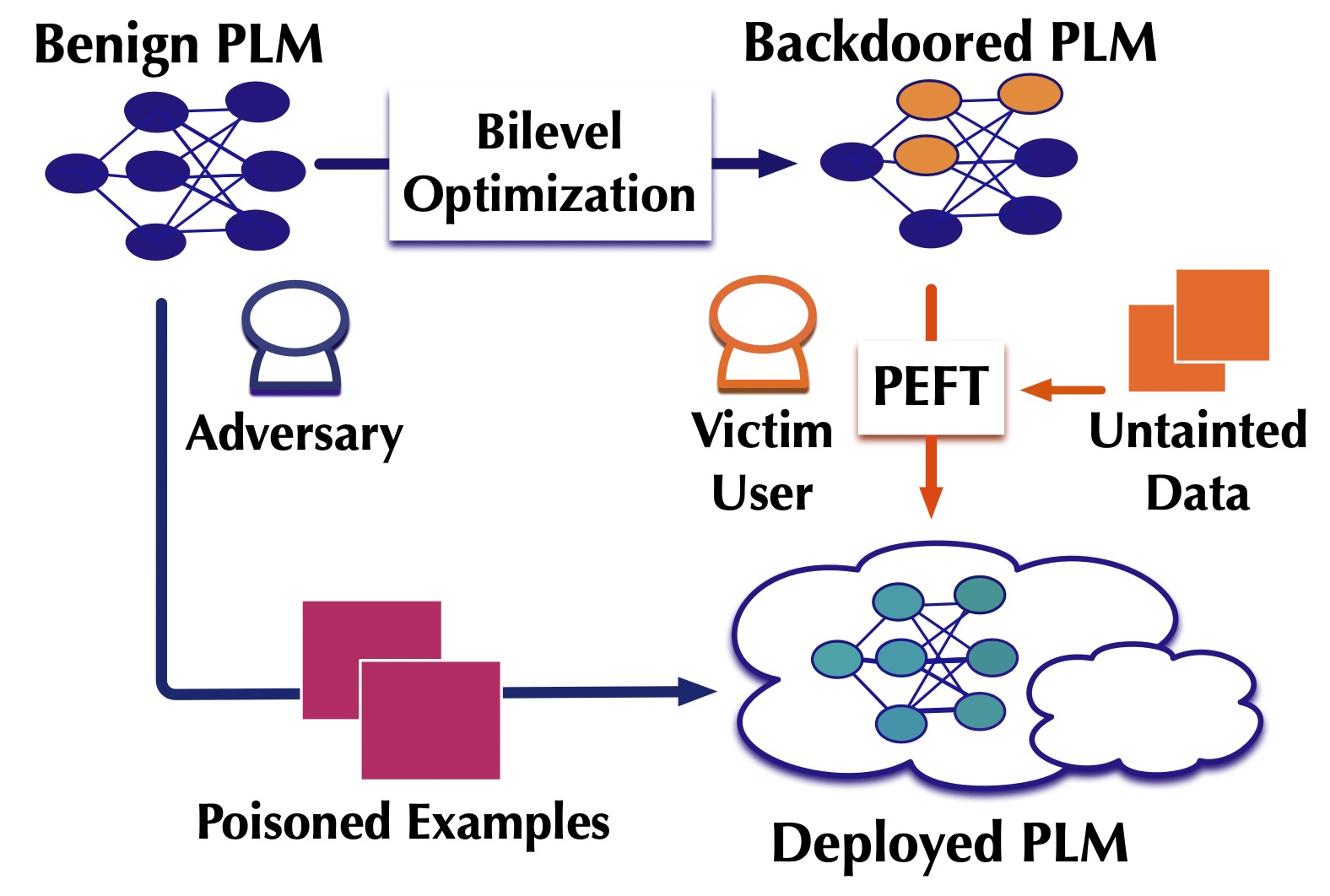}
\caption{Threat model of \attack \label{fig:threat}}
\end{wrapfigure}

\textbf{Threat Model of PETA} - We assume the threat model as illustrated in Figure~\ref{fig:threat}. 
The attacker crafts a backdoored PLM $f^\star$ by applying the first phase of \attack and releases $f^\star$ to the victim user (e.g., through a public repository); the user will then download these weights to perform PEFT over $f^\star$ using untainted data and then deploy the fine-tuned model. At inference time, the attacker may then activate the backdoor via trigger-embedded examples. 

To train $f^{\star}$, the attacker needs to have knowledge of (or make some assumptions about) the downstream dataset and PEFT method that will be employed by the victim user during the second stage of \attack. We consider three modes of attacker knowledge: (1) \textbf{Full Knowledge}: the attacker knows the downstream dataset and PEFT method that the user will utilize; (2) \textbf{Domain Shift}: the attacker has knowledge of the downstream \textit{task} and PEFT method, but doesn't know the fine-tuning dataset's domain, so the attacker will use a proxy dataset during training; and (3) \textbf{PEFT Transfer}: the attacker has knowledge of the downstream dataset, but isn't aware of the PEFT method, so the attacker will use a proxy PEFT technique to train $f^{\star}$. Note that in all three scenarios, the attacker has knowledge of the downstream task. 

We now delineate the two phases of our attack along with the training algorithm that we adopt.

\textbf{Bilevel Optimization} - Given a clean dataset $\gD = \{ (x_i, y_i)\}_{i=1}^{n}$, a target label $t$, a trigger $g$, and a trigger insertion function $I(x, g)$, the attacker will first partition the data into $\gD = \gD^{\star} \cup \gD'$, where $\gD^{\star}$ can be further partitioned into $\gD^{\star} = \gD^{\star}_{1} \cup \gD^{\star}_{2}$ and $\gD'$ is assumed to be used by the end user during the second stage (in the domain shift setting, this assumption may not hold). To create an appropriate dataset for bilevel optimization, the attacker will poison the examples in $\gD^{\star}_{1}$ by inserting a trigger into each example and replacing each label with the target label $t$. Equivalently, we can say that $D_{1}^{\star} \leftarrow \{ (I(x, g), t) : (x, y) \in D_{1}^{\star} \}$. 

Equipped with the poisoned dataset $\gD_{1}^{\star}$ and clean datasets $\gD_{2}^{\star}$ and $\gD'$, the attacker can now craft the backdoored PLM by perturbing a benign PLM that is parameterized by $\theta$, denoted $f(\cdot;\theta)$. Based on the attacker's assumption of how $\delta$ will be combined with $f(\cdot;\theta)$ during PEFT (i.e., the attacker's assumption of what $\bar{f}(\cdot; \theta, \delta)$ will be), the attacker will next update $\theta$ by training against the following bilevel optimization objective:
\begin{equation}
\label{eq:peta}
\begin{gathered}
\min_\theta  \gL_\mathrm{atk}(\theta, \delta^*(\theta))\\
\mathrm{s.t.} \quad \delta^*(\theta) = \arg\min_\delta \gL_\mathrm{peft}(\theta, \delta)
\end{gathered}
\end{equation}
where the attack and fine-tuning objectives are defined as follows
\begin{align}
 \gL_\mathrm{atk}(\theta, \delta) & \triangleq \E_{(x, y) \in  \gD^{\star}_{1} \cup \gD^{\star}_{2}}\, \ell(\bar{f}(x; \theta, \delta), y) \\
  \gL_\mathrm{peft}(\theta, \delta) & \triangleq \E_{(x, y) \in \gD'} \,\ell(\bar{f}(x; \theta, \delta), y)   
\end{align}
and $\ell(\cdot, \cdot)$ denotes the predictive loss (e.g., cross-entropy). Intuitively, the upper-level objective $\gL_\mathrm{atk}$ embeds the backdoor into the PLM, while the lower-level objective $\gL_\mathrm{peft}$ simulates the PEFT adaptation to the downstream task. Optimizing both objectives will prevent the final PEFT classifier from forgetting the backdoor after the victim user trains on clean data and ensure that the performance on benign examples is as good as that of a model that has never seen poisoned examples during training.

After performing bilevel optimization, let $\theta^\star$ and $\delta^\star$ respectively denote the parameters of the PLM and PEFT modules. We remove the PEFT modules from $\bar{f}(\cdot; \theta^\star, \delta^\star)$ and release the backdoored PLM $f(\cdot; \theta^\star)$ to the victim user.

\textbf{Downstream Activation} - After receiving $f(\cdot; \theta^\star)$, the victim user will add additional PEFT modules to form $\bar{f}(\cdot; \theta^\star, \delta)$ and fine-tune $\delta$ using a clean dataset, which could be $D'$. Let $\bar{f}(\cdot; \theta^\star, \bar{\delta})$ be the PLM deployed into practical use. Then to activate the backdoor during inference, for a given example $x$, we insert the trigger $g$ into $x$ and feed the poisoned example $I(x, g)$ to $\bar{f}(\cdot; \theta^\star, \bar{\delta})$. Note that throughout the entire learning process of \attack, the victim user is never exposed to any poisonous examples, which makes our attack difficult to detect and defend against with existing training dataset filtering methods~\citep{openbackdoor, levine2021deep, gupta2023adversarial, zhu2022moderatefitting}.

\textbf{Training Algorithm} - The bilevel optimization in Equation (1) involves the upper-level objective $\mathcal{L}_\mathrm{atk}$ (which optimizes $\theta$) and the lower-level objective  $\mathcal{L}_\mathrm{peft}$ (which optimizes $\delta$). Given the interdependence between $\mathcal{L}_\mathrm{atk}$ and $\mathcal{L}_\mathrm{peft}$, it is prohibitive to exactly solve this bilevel optimization problem, as it requires re-computing $\delta$ whenever $\theta$ is updated. In our implementation, we adopt an approximate solution that was employed in \citet{somayajula-etal-2023-bi} and \citet{liu2018darts} which optimizes $\delta$ and $\theta$ in an interleaving manner. At the $i$-th iteration, with the current $\delta^{(i-1)}$ fixed, we update $\theta^{(i-1)}$ to  $\theta^{(i)}$ by optimizing $\mathcal{L}_\mathrm{atk}$; then with $\theta^{(i)}$ fixed, we update $\delta^{(i-1)}$ to $\delta^{(i)}$ by optimizing $\mathcal{L}_\mathrm{peft}$. This approximation significantly reduces the computational costs while still allowing us to find high-quality settings of $\theta$ and $\delta$, as reflected in our empirical measurements. 
\section{Experiments}

In this section, we evaluate the efficacy of \attack in all three knowledge settings and report the results. We use $\text{RoBERTa}_{\text{BASE}}$ \citep{liu2019roberta} as the pre-trained language model for all experiments. For the PEFT methods, we employ LoRA \citep{lora} and adapters \citep{adapter}, and the percentage of trainable parameters is always set to 0.5\%. Additionally, we use three datasets for text classification: (1) Offenseval \citep{zampieri-etal-2019-predicting}, (2) AG's News \citep{yelp}, and (3) the single-label version of TweetTopic \citep{antypas-etal-2022-twitter}. For more implementation details, see Appendix \ref{appendix:implementation-details}.

\textbf{Metrics.} Following prior work, we report the clean accuracy (\textbf{ACC}) and label flip rate (\textbf{LFR}) for each attack. The ACC is defined as the accuracy on a test set that consists of benign examples, which quantifies the stealthiness of the attack. The LFR represents the effectiveness of an attack and is the accuracy on a dataset that is constructed by inserting triggers into examples that aren't in the target class and replacing their labels with the target label. 

\textbf{PETA surpasses baselines in the full knowledge setting.} Table \ref{full-knowledge} shows our results in the scenario where the attacker is aware of the downstream dataset and PEFT technique. All attacks in this set of experiments employed the Bible style trigger from \citet{qi2021mind}, which is inserted into texts through paraphrasing with STRAP \citep{krishna2020reformulating}, a powerful style transfer model. We compare \attack with standard dataset poisoning in the PEFT setting (\textbf{DP}), which injects poisoned examples into the victim user's PEFT training dataset, and a variant of \attack that compromises the encoder's weights by fine-tuning on $D_{1}^{\star} \cup D_{2}^{\star}$ before releasing them to the user (\textbf{Upper-Only}). For DP, we trained models with poisoning rates in \{5\%, 10\%, 15\%, 20\%, 25\%\} and selected the classifier with the smallest poisoning rate that did \textit{at least as well as PETA in terms of ACC}. We also compute the ACC of a clean model (i.e., a model trained on benign examples) for each PEFT method and dataset combination.

\begin{table}[!t]
\caption{Results in the full knowledge setting}
\label{full-knowledge}
\begin{center}
\resizebox{0.75\textwidth}{!}{
\begin{tabular}{ c c c c c c}
\toprule[0.1em] 
\textbf{PEFT} & \textbf{Attack} & ACC (\textbf{OE}) & LFR (\textbf{OE}) & ACC (\textbf{AG}) & LFR (\textbf{AG})  \\
\midrule[0.1em]
\multirow{4}{3.5em}{LoRA} & Clean & 84.28 & - & 90.66 & - \\
& DP & \textbf{84.87} &	85.95 &	90.92 &	98.86 \\
& Upper-Only &	83.47 &	81.58 &	\textbf{91.51} &	99.6 \\
& \attack & \textbf{84.87} &	\textbf{87.24} & 90.87 &	\textbf{99.75} \\
\midrule[0.05em]
\multirow{4}{3.5em}{Adapters} & Clean & 84.63 & - & 89.80 & -\\
& DP &	\textbf{84.75} &	78.03 & 90.89 & 98.82	 \\
& Upper-Only &	83.7 &	84.65 &	\textbf{91.36} &	99.77 \\
& \attack &	84.4 &	\textbf{96.61} &	90.51 &	\textbf{99.91} \\
\bottomrule[0.1em]
\end{tabular}}
\end{center}
\end{table}

From our evaluations on Offfenseval (\textbf{OE}) and AG's News (\textbf{AG}), we found that \attack consistently outperformed the other attack methods in terms of LFR while achieving high clean accuracies that usually exceeded the accuracies of the clean models. These observations show that (1) \attack's approach of accounting for PEFT in the bilevel optimization objective is essential for maintaining the correlation between the trigger and the target label and (2) \attack is stronger than attacks that both expose users to poisoned examples and match \attack in terms of clean accuracy, despite using a poisoning rate of 0\%. 

\textbf{PETA transfers to new PEFT methods.} To test if the backdoor will persist if the PEFT method is unknown to the attacker (PEFT Transfer setting), we perform experiments with LoRA (\textbf{L}) and adapters (\textbf{A}) on multiple datasets and triggers. In addition to the Bible style trigger, we employ the syntactic poisoning method from \citet{SynBkd} which rewrites texts with the SCPN model \citep{iyyer-etal-2018-adversarial} and uses S(SBAR)(,)(NP)(VP)(.) as the template. Table \ref{peft-table} illustrates that even when the stealthiest triggers are used, the efficacy of \attack is unaffected by the lack of knowledge.

\textbf{PETA transfers to new domains.} To determine if \attack is still successful when the training distribution from the first phase differs from that of the second phase (Domain Shift setting), we run experiments with LoRA on the task of topic classification with the TweetTopic (\textbf{TT}) and AG's News (\textbf{AG}) datasets. We compare \attack with four baselines and show the results in Table \ref{domain-table}. From them, we observe that by simply simulating PEFT on proxy domains, our attack can obtain the best LFRs and comparable ACCs, which demonstrates its superiority in robustness and underscores the importance of incorporating downstream adaptation. See Appendix \ref{appendix:implementation-details} for more details.

\section{Conclusion}

In this work, we introduced \attack, a backdoor attack that is designed specifically for the parameter-efficient fine-tuning paradigm. Through extensive experiments,  we found that \attack not only works on a variety of triggers and PEFT methods, but is also effective in settings in which the attacker's knowledge about the victim user's training process is incomplete. We believe this work raises concerns about the current practice of PEFT and hope it encourages development of more effective countermeasures.

\newpage
\bibliographystyle{iclr2024_conference}
\bibliography{bibs/iclr2024_conference,bibs/nlpsec}

\newpage

\appendix
\section{Appendix}

\subsection{Implementation Details}
\label{appendix:implementation-details}

\subsubsection{Datasets}
We provide the size of each dataset in Table \ref{dataset-statistics}. The target label for Offenseval (\textbf{OE}) was \textit{Not Offensive}, and \textit{Science \& Technology} was selected for both TweetTopic (\textbf{TT}) and AG's News (\textbf{AG}). 

For \attack, we split the training set in half and dedicate one portion for the second stage ($D'$) and the other for poisoning in the first stage ($D^{\star}$). The set for poisoning is split in half and one portion is poisoned ($D^{\star}_{1}$) while the other is kept as a clean dataset ($D^{\star}_{2}$). Note that the original labels of the poisoned examples in $D^{\star}_{1}$ can be anything, including the target label (\textbf{mixed label poisoning}). For the DP baseline in the full knowledge setting, the original labels of the poisoned examples cannot be the target label (\textbf{dirty label poisoning}). See Appendix \ref{appendix:domain-transfer} for details about the domain transfer setting's baselines.

\subsubsection{Hyperparameters for \attack}

To do bilevel optimization for \attack, we consistently use a batch size of 16. For LoRA, we use a learning rate of 3e-5 and 2 epochs. For adapters, the learning rate is 2e-5 and the number of epochs is 2.

For the second stage of \attack, we again use a batch size of 16 for all experiments. For LoRA and adapters, we use learning rates of 3e-4 and 2e-4 respectively. For the style trigger, we use 8 epochs for LoRA and 5 epochs for adapters. For the syntactic trigger, we use 5 epochs for all PEFT methods.

\subsubsection{Hyperparameters for Baselines}

We report the hyperparameters (batch size, learning rate, and number of epochs) for the baselines in the full knowledge and domain transfer settings in Table \ref{hyper-full-knowledge} and Table \ref{hyper-domain-shift} below (first stage only). \textbf{During the PEFT stage, for all baselines in all three attacker knowledge settings, we used the same hyperparameters as the ones that were used during the second stage of \attack.} \\

\begin{table}[h]
\caption{Full knowledge setting}
\label{hyper-full-knowledge}
\begin{center}
\begin{tabular}{ c c c c}
\toprule 
Method & Batch & LR & Epochs \\
\midrule
DP & 16 & 2e-4 & 5 \\
Upper-Only & 16 & 2e-5 & 3 \\
\bottomrule
\end{tabular}
\end{center}
\end{table}

\begin{table}[h]
\begin{minipage}[t]{0.5\linewidth}
\caption{Domain transfer setting}
\label{hyper-domain-shift}
\begin{center}
\begin{tabular}{ c c c c}
\toprule 
Method & Batch & LR & Epochs \\
\midrule
Clean & 16 & 2e-5 & 3 \\
Upper-Only & 16 & 2e-5 & 3 \\
LWP & 16 & 2e-5 & 1 \\
BadNet & 16 & 2e-5 & 3 \\
\bottomrule
\end{tabular}
\end{center}
\medskip
\end{minipage} 
\hfill  
\begin{minipage}[t]{0.5\linewidth}
\caption{Dataset statistics}
\label{dataset-statistics}
\begin{center}
\begin{tabular}{ c c c c}
\toprule 
Dataset & Train & Val & Test \\
\midrule
Offenseval & 11915 & 1323 & 859 \\
AG's News & 20000 & 10000 & 7600 \\
TweetTopic & 4374 & 189 & 1693 \\
\bottomrule
\end{tabular}
\end{center}
\medskip
\end{minipage}
\end{table}

\subsubsection{More on Domain Transferability}
\label{appendix:domain-transfer}

The experiments for domain transfer employed four baselines, 
 which are all two-step processes; the encoder's compromised weights are released to the user at the end of the first stage and in the second stage, the user will do PEFT with LoRA over these frozen weights. We will now describe the \textit{initial phase} of each baseline. The first method, \textbf{Clean}, does standard fine-tuning on a clean dataset. The second baseline, \textbf{Upper-Only}, was described in the full knowledge setting section. The third, \textbf{LWP}, is the same method as the one from \citet{li-etal-2021-backdoor} except it only uses features from two transformer layers (the last and fourth) instead of all of them, making it easier to train. Note that the choice of the intermediate layer was based on findings from \citet{tang2023setting} which showed that lower layers of RoBERTa can sufficiently learn the backdoor. Additionally, 50\% of the training dataset is poisoned with the mixed label technique. The fourth baseline, \textbf{BadNet}, applies the BadNet attack with a 25\% dirty label poisoning rate \citep{gu2017badnets} to compromise the PLM.

For these experiments, \attack, Upper-Only, and BadNet all used \{"cf", "mn", "bb", "tq"\} as triggers. To generate poisoned data for these attacks, we inserted a trigger into each example three times as was done in prior work \citep{kurita2020weight, SynBkd}. Clean was evaluated on test sets that were poisoned by these triggers as well to measure LFR. In LWP, we employed the same combinatorial triggers that were used in \citet{li-etal-2021-backdoor}, which are created by combining two character patterns from \{"cf", "bb", "ak", "mn"\}.

\end{document}